\documentclass[final,5p,times,twocolumn]{elsarticle}

\usepackage{amsmath} 
\usepackage{color}
\usepackage{makecell}
\usepackage{balance} 

\usepackage{arydshln}
\usepackage{booktabs}
\usepackage{adjustbox}
\usepackage[ruled, linesnumbered]{algorithm2e}
\usepackage{bbm}
\usepackage{bm}
\usepackage{framed} 
\usepackage{graphicx}
\usepackage{graphics}
\usepackage{multirow}
\usepackage{float}
\usepackage{microtype}      
\usepackage{rotating}
\usepackage{subcaption}

\usepackage{breakcites}
\usepackage[toc,page]{appendix}

\usepackage{hyperref}

\usepackage{amsthm} 

\usepackage[capitalize]{cleveref}
\crefname{section}{Sec.}{Secs.}
\Crefname{section}{Section}{Sections}
\Crefname{table}{Table}{Tables}
\crefname{table}{Tab.}{Tabs.}

\AtBeginDocument{%
  \providecommand\BibTeX{{%
    \normalfont B\kern-0.5em{\scshape i\kern-0.25em b}\kern-0.8em\TeX}}}

\usepackage{adjustbox}
\usepackage{arydshln} 
\usepackage{xcolor} 
\definecolor{blue2}{HTML}{3399FF}
\definecolor{green2}{HTML}{82B366}
\definecolor{orange2}{HTML}{FFCE9F}
\def\doubleunderline#1{\underline{\underline{#1}}}

\usepackage{lineno}

\journal{Elsevier}

\begin{document}
\date{}
\begin{frontmatter}



\title{Occlusion-Robust FAU Recognition by Mining \\
Latent Space of Masked Autoencoders}


\author[contact1]{Minyang Jiang}
\ead{minyang@ece.ubc.ca} 
\author[contact2]{Yongwei Wang\corref{cor1}}
\ead{yongweiw@ece.ubc.ca} 
\fntext[1]{Y. Wang did part of the work at UBC, and he has moved to NTU.}

\author[contact3] {Martin J. McKeown}
\ead{martin.mckeown@ubc.ca} 
\author[contact1]{Z. Jane Wang}
\ead{zjanew@ece.ubc.ca} 

\address[contact1]{Department of Electrical and Computer Engineering, University of British Columbia, Vancouver, BC, Canada}
\address[contact2]{Joint NTU-WeBank Research Center on Fintech, Nanyang Technological University, Singapore} 
\address[contact3]{Department of Medicine, University of British Columbia, Vancouver, BC, Canada}

\cortext[cor1]{Corresponding author.}

\begin{abstract}
Facial action units (FAUs) are critical for fine-grained facial expression analysis. Although FAU detection has been actively studied using ideally high quality images, it was not thoroughly studied under heavily occluded conditions. In this paper, we propose the first occlusion-robust FAU recognition method to maintain FAU detection performance under heavy occlusions. Our novel approach takes advantage of rich information from the latent space of masked autoencoder (MAE) and transforms it into FAU features. Bypassing the occlusion reconstruction step, our model efficiently extracts FAU features of occluded faces by mining the latent space of a pretrained masked autoencoder. Both node and edge-level knowledge distillation are also employed to guide our model to find a mapping between latent space vectors and FAU features. Facial occlusion conditions, including random small patches and large blocks, are thoroughly studied. Experimental results on BP4D and DISFA datasets show that our method can achieve state-of-the-art performances under the studied facial occlusion, significantly outperforming existing baseline methods. In particular, even under heavy occlusion, the proposed method can achieve comparable performance as state-of-the-art methods under normal conditions.

\end{abstract}


\end{frontmatter}

\section{Introduction}
\label{sec:intro}

The Facial Action Coding System (FACS) \cite{FACS1} is a comprehensive system that breaks down facial expressions into individual components of muscle movement, which are called Action Units (AUs). It is widely adopted to describe fine-grained facial behaviors. Automatic action unit detection enables efficient facial analysis and can be used in a wide range of applications including security, clinic, entertainment, and education \cite{fau_survey_2020}. 

With the recent advancement of deep neural networks (DNNs) and high-quality image datasets, the performance of computer vision tasks has been improved tremendously including facial action unit (FAU) detection. Some pioneering studies (e.g., \cite{zhao2015joint, jaiswal2016deep, li2017eac, eacnet2017, corneanu2018deep}) take advantage of DNNs to extract local and global facial appearance features, and they have remarkably improved the accuracy of FAU detection over traditional approaches using hand-crafted features \cite{cohn1999automated, whitehill2006haar}. More recently, works \cite{liu2020relation, shao2018deep} further improve the detection performance by combining the appearance features with domain knowledge of dependencies between AUs. To further capture these AU dependencies automatically, Luo et al. \cite{luo2022learning} propose ME-GraphAU, a node and edge feature learning approach that can achieve state-of-the-art performance in FAU recognition. Despite the promising performance of these methods, they all rely on high-quality images and videos gathered from well-controlled lab environments with full facial region and action units exposed. As a result, directly applying these methods often suffers from significant performance degradations in the presence of occlusion, particularly for heavily occluded face images. Indeed, even for ME-GraphAU \cite{luo2022learning}, the state-of-the-art approach, our preliminary studies show that its F1-score drops sharply from 65.5\% to 30.7\% by randomly masking 50\% of facial regions.

In many real-world scenes, the captured face images can be partially or even heavily occluded. Thus, the inference on occluded facial images has been a long-standing problem in face recognition-related tasks, e.g., occlusion-roust face identification and recognition \cite{song2019occlusion,zhang2022learning,qiu2021end2end,zeng2021survey}. However, to our best knowledge, there have been no explorations yet specifically developed for the occluded FAU recognition task. Facial occlusion is often characterized as an intractable problem, thus it presents particular challenges to existing occlusion-unaware methods \cite{zeng2021survey}. First, a large portion of data that contains discriminative appearance features may be missing, which leads to severe performance degradation. Earlier region-based AU prediction models that only infer from local features will fail because of missing information on occluded regions \cite{li2017eac}. Some other typical models considering dependencies among AUs by graph models will also encounter the problem of missing node features \cite{liu2020relation, song2021uncertain, luo2022learning}. Second, we often do not have prior information about the occluded regions (e.g., position, shape), further increasing the difficulty of producing accurate FAU predictions.     

Recent studies in compressed sensing theory reveal an intriguing phenomenon that, image signals contain much redundant information such that missing image regions may be recovered with high probability under proper sampling conditions \cite{rani2018systematic,kulkarni2016reconnet, wang2018revhashnet}. Moreover, different facial AUs often mutually influence each other \cite{luo2022learning}, thus the activation status of one missing AU may be inferred from neighboring AUs. Based on these observations, as the first attempt, we aim to exploit and reconstruct missing AUs from occluded facial images prior to FAU recognition. To show the feasibility of the concept, we explore the adoption of masked autoencoder (MAE) structure \cite{he2022masked} for image reconstruction, which achieves state-of-the-art performance in the self-supervised learning regime. In particular, MAE has been demonstrated to well recover an image even with 75\% randomly masked missing regions. Despite the effectiveness of MAE in reconstructing natural images, there are two key issues that may hinder the direct employment of MAE for occluded FAU recognition. First, the decoding process of MAE requires the location information of occluded regions as \textit{a priori}; while usually we don't have such information or need extra efforts to obtain it in a real test scene. Moreover, the decoding network in MAE also causes a large computational overhead. 

To address these challenges above, we propose a simple yet effective and efficient framework based on off-the-shelf masked autoencoders. In our preliminary study, we leverage a pretrained MAE to predict occluded missing facial regions and observe surprisingly good overall reconstruction quality. This key observation indicates that the bottleneck layer of a pretrained MAE is capable of capturing essential knowledge of relations between different action units, thus well recovering missing facial action units. We are then motivated to mine discriminative feature information from the latent space of the MAE. Meanwhile, we can bypass the redundant decoding process to be much more efficient. To make the learning process more effective, we propose to perform node and edge knowledge distillation simultaneously to further aid the model find the mapping between the latent space vector of MAE and features needed for FAU recognition. The superior performance is validated through experiments on benchmark datasets under different facial occlusion conditions.

The contributions of this paper can be summarized as follows:
\begin{enumerate}
    \item As the first attempt, we specifically explore the facial action unit recognition task and investigate its feasibility under heavily occluded conditions.  
    
    \item We propose a novel and effective reconstruction-based FAU recognition approach by mining the latent space of off-the-shelf masked autoencoders.  
    
    \item We further improve the efficiency of the occlusion-insensitive model by transferring the latent space feature of the masked autoencoder to FAU features.
    
    \item We perform experiments on two benchmark datasets and demonstrate that our method can achieve comparable performances with occlusion-free images even for 50\% heavily occluded facial images. 
 
\end{enumerate}

\begin{figure*}[!htbp]
  \centering
  \includegraphics[width=1.0\linewidth]{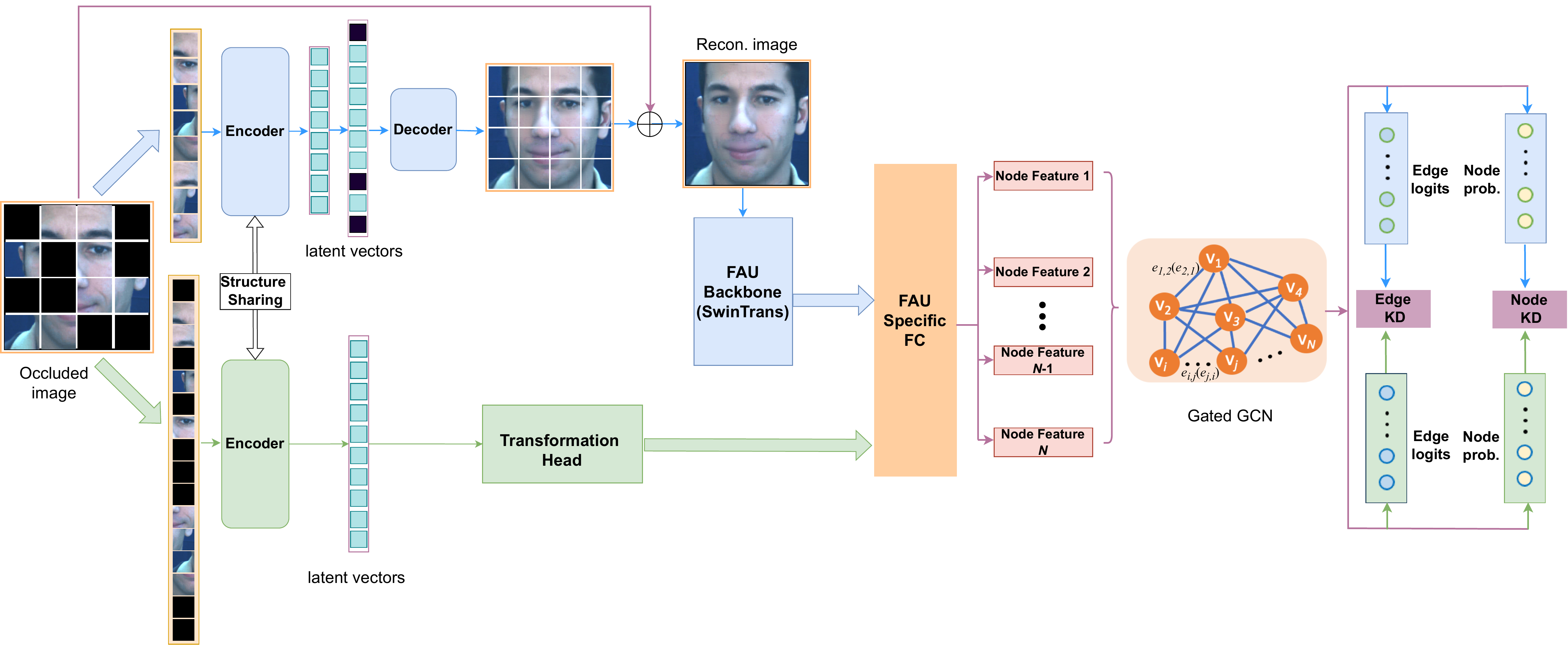}
  \caption{
  Overview of the proposed framework for occlusion-robust FAU recognition. The {{\color[HTML]{3399FF} blue}} path on top is the reconstruction-based teacher model (Section \ref{sec:recons_module}). The {{\color[HTML]{82B366} green}} path on the bottom is the student network learning through knowledge distillation (Section \ref{sec:kd}). The {{\color[HTML]{FFCE9F} orange}} part in the middle represents the GCN structure shared by both the teacher and student models, conducting AU feature and edge relationship learning (Section \ref{sec:fau_detection}).}
  \label{fig:framework}
\end{figure*}

\section{Related Work}
\label{sec:related}
In this part, we will briefly review existing works that are closely related to our proposed approach, including deep learning-based FAU detection models, masked autoencoders, and knowledge distillation techniques.

\subsection{Facial Action Unit Detection}
Early works treat action unit detection as a patch-learning problem where detected landmarks define the region of interest. In JPML \cite{zhao2015joint}, joint patch and multi-label learning are introduced where a discriminative subset of patches are used to identify target AUs. The authors further improved their method in DRML \cite{zhao2016deep} by combining deep region and multi-label learning into a unified deep network using a specifically designed region layer to replace conventional convolutional layers. The region layer can capture the local appearance change of different facial regions. In EAC-Net \cite{eacnet2017}, authors design a fixed attention map based on facial landmarks to enhance the AU feature learning in regions of interest (ROI). JAA-Net \cite{shao2018deep} jointly estimates the location of landmarks and the presence of action units. In this work, the adaptive attention map for each action unit is computed separately using estimated landmarks, yielding precise local features. Work \cite{jacob2021facial} uses the ROI attention module to predict attention maps directly using the supervision from landmarks. SEV-Net \cite{yang2021exploiting} combines the embeddings of semantic description of AUs with visual features to generate a cross-modality attention map, assisting the model to learn discriminative features from meaningful regions.

Besides learning better local features, the focus of FAU detection gradually shifts towards AU relationship modeling. DSIN \cite{corneanu2018deep} uses a recurrent neural network to perform structure inference on fused local and global features. The authors propose an iterative structure inference process to simulate the fully connected graph which captures the relationship between AUs. AU-GCN \cite{liu2020relation} proposes a graph convolutional network-based framework for modeling AU relationships. Individual AU features are fed into a GCN as nodes, and a fixed connection matrix is constructed based on statistical results on each training set. In the work of UGN \cite{song2021uncertain}, a probabilistic mask is used on graph edges to simultaneously capture dependencies and underlying uncertain information among AUs. The uncertainties are also used to select hard samples to improve training efficiency. Unlike the previous GCN-based methods where the adjacency matrix only represents the connectivity between nodes, ME-GraphAU \cite{luo2022learning} employs edge feature learning where a pair of multi-dimensional edge features are learned between each pair of AUs. The combination of node and edge features captures both the activation status of each AU and the association between them. This method extracts reliable task-specific relationship cues for AU recognition and achieved state-of-the-art results on two widely used AU datasets. Although learning rich node and edge features, ME-GraphAU still relies on the visibility of the full facial region where heavy occlusion causes significant degeneration in performance. This work improves ME-GraphAU in the presence of heavy occlusions.

\subsection{Masked Image Modeling}
Performing computer vision tasks on masked images is called masked image modeling (MIM). Models can learn meaningful representations by reconstructing masked images, and it is promising to apply MIM for self-supervised pre-training. BEiT \cite{bao2022beit} is one of the first works to use masked image modeling tasks to pretrain vision transformers where the goal is to recover original visual tokens from a masked image. SimMIM \cite{Xie_2022_CVPR} proposes a simple framework to demonstrate that masked image modeling provides the model superior representation-learning performance by experiments on large-size masked patches, simple pixel-wise regression, and lightweight prediction heads. MAE \cite{he2022masked} adopts an asymmetric encoder-decoder architecture to produce informative latent representation by training image reconstruction tasks on 75\% masked images. MAE can learn models with high capacities that generalize well, not only producing high-quality image reconstruction but also improving the performance of downstream learning tasks. Here we consider heavy occlusion as a masked image modeling task where representational features learned from non-occluded parts should benefit from reconstructing the occlusion portion and provide meaningful information to the FAU detection task.

\subsection{Knowledge Distillation}
In a real-world deployment, it is desirable that an FAU recognition model is lightweight, to be resource-efficient. However, a smaller FAU model is often associated with performance degradation. Therefore, we aim to leverage knowledge distillation (KD) to create a lightweight FAU model while preserving high accuracy. 

KD is a popular model compression method where smaller models learn from models with higher knowledge capacity \cite{gou2021knowledge,ding2023distilling}. KD is commonly used in multi-class classification tasks. Different KD methods have been proposed, such as logit-based KD \cite{hinton2015distilling}, feature-based KD \cite{romero2014fitnets}, self-supervision signals guided KD, etc \cite{xu2020knowledge, wang2022ssd}. KD is also commonly used in multi-label classification tasks to simplify large models size or improve performances through distilled knowledge. \cite{liu2018multi} builds an efficient multi-label image classification model by distilling knowledge from a weakly supervised detection model. CPSD \cite{xu2022boosting} boosts the performance of multi-label image classification through self-distillation. \cite{song2021handling} proposes uncertainty distillation to address the problem of hard samples in multi-label image classification. In this work, we intend to perform KD for both nodes and edge features, where we formulate them as multi-label KD and multi-class KD, respectively.

\section{Methodology}
\label{sec:method}
In this section, we present the framework of the proposed method as shown in Fig.~\ref{fig:framework}. The overall framework consists of three modules: the MAE reconstruction module, the FAU detection module, and the knowledge distillation module. We elaborate on each component individually as follows.

\subsection{MAE Reconstruction Module}
\label{sec:recons_module}
The MAE reconstruction module is rooted in the idea of the masked autoencoder (MAE), a state-of-the-art masked imaging modeling method where the original image can be reconstructed by observing only partial signals. More specifically, the encoder MAE firstly maps observed patches into a latent space representation. Then, empty latent space vectors representing masked patches are added at corresponding positions in the latent representation and then projected back to the image space through a decoder. Finally, the pixel values of masked patches are reconstructed in the encoding and decoding process. In this module, there are mainly four components: masking, patch encoding, latent space representation decoding, and image reconstruction.

The masking process is to simulate random occlusions on facial images.  MAE is a vision transform (ViT) \cite{vit} based method that operates on image tokens, where images are divided into non-overlapping patches. To better simulate real-world random occlusions, we consider two types of patch-based masking occlusions: random small-patch masking and large-block masking. The former type of masking is similar to conventional MAE settings where a random subset of small patches is selected. This type of masking strategy is to simulate random and small-patch occlusions on faces (e.g., hair, sunglasses, fingers). The latter masking strategy is used to simulate large-block occlusion regions (e.g., covered by facial masks or palms). In this case, a random large-block region consisting of many patches is chosen to be masked. Such large block occlusion further increases the difficulty of both image reconstructions and FAU detection since only signals from distant patches are available.

The encoder component of the reconstruction module only takes in unmasked patches. The encoder follows standard ViT \cite{vit} operations to obtain a latent space representation: applying a linear projection to construct patch embedding, incorporating positional embedding to provide the position information of each image patch, and passing through a series of transformer blocks. Taking advantage of self-generated masks with known positions, the encoder by design can only operate on visible patches saving a large amount of computation and memory.

A decoder component is used to map the latent space representation back to the image space including the reconstruction of previously masked patches. Different from the large encoder that operates only on unmasked patches, the smaller decoder takes both encoded visible patches and learned mask tokens as input. A shared vector representing mask token is filled at each position where the missing patch needs to be predicted. After filling tokens to the set, additional positional embedding is added on all tokens to provide necessary location information, especially for mask tokens. A smaller amount of transformer blocks is used in the decoder to save computation.

As for the image reconstruction step, an MAE directly reconstructs the pixel values of masked patches. MAE uses the pixel-wise $L_2$ metric to measure the quality of reconstruction on masked image patches, and the loss can be written as, 
\begin{equation} L_{recons} =\frac{1}{N_{M}} \sum_{i \in M} \Big \|\hat{x}_i - x_i \Big \|_2 \end{equation} 
where $M$ represents the masked pixels set, $N_{M}$ is the number of masked pixels, $\hat{x}$ and $x$ corresponds to predicted pixel values and original pixel values, respectively.

However, the reconstructed image suffers from significant block artifacts due to unconstrained visible patch reconstruction. To reduce the noise caused by such artifacts on downstream FAU tasks, we again use the positional information of masks to combine the visible patches from the original image with reconstructed masked patches and form a better-quality facial image for the next FAU detection module.

\subsection{FAU Detection Module}
\label{sec:fau_detection}
This module is a self-contained facial action unit detection where the inputs are regular RGB images and the outputs are the activation probability of each facial action unit. To guarantee a good FAU detection performance, this module is based on the state-of-the-art FAU detection method \cite{luo2022learning}. The detection module can be further split into AU feature generation and graph learning components.

To generate AU features, the face feature map $F\in \mathbb{R}^{H \times W \times C}$ ($H$, $W$, and $C$ correspond to the height, width, and channels of the feature map) is extracted by standard computer vision backbones. $N$ different fully connected layers are used on $N$ AUs respectively to selectively extract features that are specific to each AU from the full-face feature map. Global average pooling is used on each AU-specific feature map to generate $N$ feature vectors $v_i \in \mathbb{R}^C$ as AU-specific representations. 

To better model the relationships between AUs, we adopt a graph neural network approach incorporating both node and edge feature learning. This is a two-stage learning method. Node feature learning is conducted in the first stage where the learning target is to produce node features containing both the AU activation status and associations with each other on each facial display. In this stage, each AU representation is used as graph node features and the similarity between these node features determines the connectivity among nodes. More specifically, $s_{i,j} = \text{Sim}(v_i, v_j)$ and $a_{i, j} =1, \ \textrm{if} \ s_{i,j} \in \textrm{Top}_K(s_i)$, where $a_{i,j}$ represents the connection between node $i$ and $j$ in adjacency graph $A$, and $K$ represents the out-degree of each node to their closest neighbors. After the construction of nodes and edges, one GCN layer is used to update the AUs' activation status by fusing information from most related AUs. The updated AU representations can be written as,

\begin{equation}
    V^{new} = \sigma \Big (V + BN(A \cdot g_1(V)+ g_2(V)) \Big)
\end{equation}
where $V \triangleq \{v_i\in \mathbb{R}^C\}_{i=1}^N$, \ $\sigma(\cdot)$ is a non-linear activation function, $BN(\cdot)$ represents the batch normalization function, and $g_1,g_2$ denote linear layers with weight and bias. To provide a probabilistic prediction of the activation status of each action unit, a similarity calculation strategy is used here, where cosine similarity is computed between a trainable vector $t_i \in \mathbb{R}^C$ and an updated representation vector $v^{new}_i \in \mathbb{R}^C$. The vector $t_i$ is trained to be a representation of the active status of the $i-$th AU. The probability of the $i-$th AU being activated can be written as,
\begin{equation}
    p_i = S_c \Big (\sigma(t_i), \sigma(v^{new}_i) \Big )
\end{equation}
where $S_c(\cdot)$ denotes a function that computes the cosine similarity between two vectors. 

To supervise the learning of node features, a multi-label classification loss is adopted. However, there are two significant label imbalance issues due to the nature of FAU dataset collection process: First, the negative label dominates on each AU; Second, the occurrence frequency of each AU is dramatically different. To address these issues, we adopted the weighted asymmetric loss proposed in ME-GraphAU and added additional degrees of freedom to compensate for the noisiness caused by image reconstruction. We first applied asymmetric probability shifting \cite{ben2020asymmetric} on the estimated probability $p^m_i$,
\begin{equation}
    p^m_i = \max(p_i - m, 0)
\end{equation}
where $m$ denotes a margin to discard low-probability negative samples. This strategy helps reject mislabeled negative samples generated in the image reconstruction process. The AU loss now can be written as, 
\begin{equation}
    L_{AU} = -\frac{1}{N}\sum_{i=1}^Nw_i \Big [y_i\log(p^m_i)+(1 - y_i)(p^m_i)^\gamma \log(1 - p^m_i) \Big]
\end{equation}
where $w_i=\frac{N(1/r_i)}{\sum_{j=1}^N(1/r_j)}$ is pre-generated by occurrence rate $r_i$ of the $i$-th AU in the training dataset, $y_i$ is the ground truth binary label, and $\gamma$ is the hyperparameter only applies to negative sample to adjust the contribution from easy negative samples.
In this stage, only AU nodes with similar feature representations are connected, which forces the model to extract AU features containing both activation and association information. 

The second stage builds on top of the AU features learned in the first stage. In addition to associations encoded in node features, this stage aims for learning edge features that describe fine-grained relationships between AUs through additional supervision. Edge features contain much richer information than binary connectivity in the adjacency matrix.  Far away nodes in terms of similarity can still have critical relationship information contributing to the detecting activation of AUs. To acquire meaningful edge features, the model conducts two cross-attention operations,
\begin{equation}
    \text{Cross-Attention}(A, B) = \text{softmax}\Big (\frac{AW_q(BW_k)^T}{\sqrt{d_k}} \Big ) BW_v
\end{equation}
where $W_q, W_k, W_v$ are learned weights that apply a linear transformation on the query, key, and value in the attention mechanism, and $d_k$ is a scaling factor that is equal to the number of channels in $BW_k$ term. Firstly, a cross-attention operation is conducted between each AU-specific feature map and the full-face feature map, acquiring the AU activation status in terms of global face feature representation,
 \begin{equation}
     F^{face}_i=\text{Cross-Attention}_1(F^{AU}_i, F^{face})
 \end{equation}
Then, between each pair of AUs, another cross-attention operation is used to extract features that are related to both AUs,
\begin{equation}
    F^{rel}_{i,j} = \text{Cross-Attention}_2(F^{face}_i, F^{face}_j)
\end{equation}

With global average pooling on the above features map describing the relationship between pair of AUs, the edge feature vectors $E \triangleq \{e_{i,j}\in \mathbb{R}^C\}_{i,j=1}^N$ is obtained. Thus, we can form a graph $G^0=(V^0, E^0)$ containing both node and edge features. Multiple layers of GatedGCN \cite{bresson2017residual} are used on the graph to allow information propagation between AUs leading to more accurate AU activation status and richer edge features. The activation probability is generated using the similarity calculation strategy as in the node feature learning in the first stage. To further guide edge feature learning, one additional classification head is added to the final edge features. The classification head classifies 1 of the 4 possible activation status combinations of two AUs that the edge connects to. Categorical cross-entropy loss is used for edge classification, and it can be written as,
\begin{equation}
    L_{E} =\frac{1}{|E|}\sum_{i=1}^{N}\sum_{j=1}^{N}\text{CCE}\Big (y^e_{i,j}, \text{softmax}(z_{i,j}) \Big )
\end{equation} 
where $\text{CCE}(\cdot)$ is the categorical cross-entropy function, $y^e_{i,j}\in \mathbb{R}^4$ is a one-hot vector indicating 1 of the 4 co-occurrence patterns of the edge between the $i$-th and $j$-th nodes, $z_{i,j}\in \mathbb{R}^4$ denote the logits output from the edge classification head. In the second stage, the model focuses on node and edge feature learning with an MAE reconstruction module. The loss in this stage can be written as,
\begin{equation}
    L_{stage2} = L_{AU} + \lambda L_{E}
\end{equation}
where $\lambda$ is a hyperparameter that adjusts the importance of edge classification results.

\subsection{Student Module}
\label{sec:kd}
Using the above two modules, we now have a complete occlusion-robust FAU detection pipeline that can be trained end to end by learning from regular images with generated masks. By using reconstructed images from the MAE reconstruction module as input, the FAU detection module can estimate the activation status of the heavily occluded face. The purpose of reconstructing the masked facial image in RGB space is to provide supervision on creating high-quality FAU-aware face reconstruction, and reconstructed images allow maximum flexibility in terms of FAU detection model selection. However, because of the process of facial image reconstruction and FAU feature extraction, the complexity of the model greatly increases. Besides, during the testing phase, the reconstruction process requires explicit knowledge of occluded positions which is often unknown in practice. Furthermore, the square artifacts in the reconstructed facial images can potentially cause error propagation in the downstream FAU detection tasks. In the next module, we want to maintain the effectiveness of this occlusion-robust FAU detection pipeline (the teacher model) while addressing the problems by introducing feature alignment and knowledge distillation.

Since the latent space representation in MAE is capable of reconstructing masked patches of arbitrary images, it should contain generic information extracted from visible patches. Meanwhile, in the previous pipeline, FAU-related features are then extracted from the reconstructed image. This indicates that the FAU-specific features can be derived directly from the generic MAE latent space features without reconstructing the missing patches. In standard MAE training, masks are generated during the forward propagation process and the position information of masks is used in several places including selecting visible patches before the encoder, adding masked tokens in the corresponding place in the latent space representation, and combining visible patches with reconstructed ones to form better quality images. In the proposed student network, to better simulate random occlusions in a realistic scenario, we intentionally avoid the use of position information of occlusion. In the student network, tokens of all patches are fed into the encoder with additional positional embeddings indicating the location of patches. This produces the latent space representation $F^{mae}$ with the same dimensionality as standard MAE latent space representation after adding mask tokens. To mitigate the gap between MAE latent space representation and the FAU face feature map, we add a simple feature alignment component between these two feature spaces. The proposed feature alignment component uses a downsampling layer and multiple fully connected layers to selectively project MAE latent space representation into features that are significant for FAU detection. 

With features projected into the same space as features generated in the FAU detection module, the AU-specific feature generation and graph learning components from the original pipeline can be seamlessly adopted. Nevertheless, there is a large gap in terms of knowledge capacity between the model with reconstruction and the student model directly projecting features from the latent space of MAE. To efficiently transfer the rich knowledge of the well-trained large model to this compact model, we apply knowledge distillation losses on both FAU detection and edge classification targets. 

FAU detection is modeled as a multi-label classification problem where the ground truth label indicates the occurrence of certain AU. The output of the teacher model contains probabilistic estimations of the occurrence which provides meaningful likelihood information that binary ground truth labels do not have. To incorporate probability information learned from the teacher model, we minimize the per AU Kullback-Leibler (KL) divergence between the output from the teacher and student model. The AU distillation loss can be written as,
\begin{equation}
    L^{AU}_{kd} = \frac{1}{N}\sum_{i=1}^N D_{\text{KL}}(p^s_i, p^t_i) + D_{\text{KL}}(1-p^s_i, 1-p^t_i)
\end{equation}
where $D_{KL}(\cdot)$ denotes the KL divergence function and $p^s_i, p^t_i$ are activation probability outputs for the $i$-th AU from the student and teacher models, respectively.

The edge features are also critical in the detection algorithm, so a 4-class classification problem is set up to guide the model to learn representative association features between AUs. To efficiently transfer knowledge in terms of edge features, we adopt a typical knowledge distillation on the logit layer output of the edge classification head \cite{hinton2015distilling}, the edge distillation loss can be written as,
\begin{equation}
    L^{edge}_{kd} = T^2D_{KL} \Big (\text{softmax}(z^s/T), \text{softmax}(z^t /T) \Big )
\end{equation}
where $T$ is the temperature hyperparameter that adjusts the smoothness of probability, and $z^s, z^t$ are logit output from the edge classification head of student and teacher models,  respectively.

The overall knowledge distillation loss can be written as,
\begin{equation}
    L_{kd} = L^{AU}_{kd} + \beta L^{edge}_{kd}
\end{equation}
where $\beta$ are hyperparameters to adjust the relative weight between AU distillation loss and edge distillation loss.

Finally, we have our overall training loss for the student network by combining with the AU detection loss, edge classification loss, and knowledge distillation loss,
\begin{equation}
    \textrm{Loss} = L_{AU} + \lambda L_{E} + \alpha L_{kd}
\end{equation}

\section{Experiments}
\label{sec:experinment}
In this section, we will empirically demonstrate the effectiveness of the proposed FAU recognition method in the presence of different occlusion conditions. We first describe our experimental setup (e.g., datasets, metrics) and then present comparison results with state-of-the-art methods on two benchmark datasets. Experimental results show that the proposed method significantly outperforms state-of-the-art methods in the presence of heavy occlusions. 

\begin{table*}[]
\centering
\begin{tabular}{|c|c|c|c|c|}
\hline
\raisebox{-.5\height}{\includegraphics[scale=0.25]{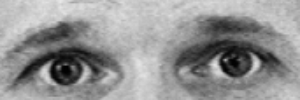}} & \raisebox{-.5\height}{\includegraphics[scale=0.25]{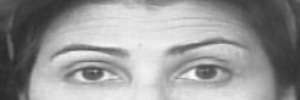}} & \raisebox{-.5\height}{\includegraphics[scale=0.25]{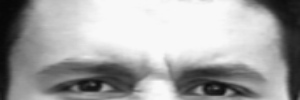}} & \raisebox{-.5\height}{\includegraphics[scale=0.25]{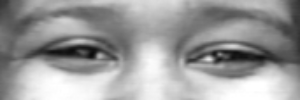}} & \raisebox{-.5\height}{\includegraphics[scale=0.25]{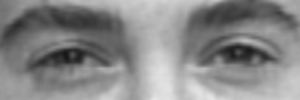}} \\ 
\thead{AU1: Inner\\ Brow Raiser}                                           & \thead{AU2: Outer\\ Brow Raiser} & \thead{AU4: \\Brow Lowerer} & \thead{AU6:\\ Cheek Raiser} & \thead{AU7:\\ Lid Tightener} \\ \hline
\raisebox{-.5\height}{\includegraphics[scale=0.25]{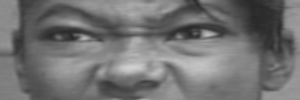}}                                                                 & \raisebox{-.5\height}{\includegraphics[scale=0.25]{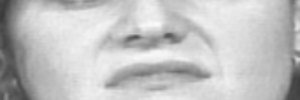}} & \raisebox{-.5\height}{\includegraphics[scale=0.25]{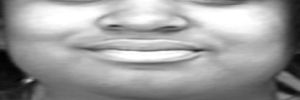}} & \raisebox{-.5\height}{\includegraphics[scale=0.25]{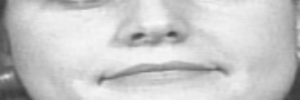}} & \raisebox{-.5\height}{\includegraphics[scale=0.25]{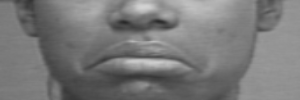}} \\ 
\thead{AU9:\\ Nose Wrinkler}                                 &      \thead{AU10: Upper\\ Lip Raiser}                          & \thead{AU12: Lip\\ Corner Puller}  & \thead{AU14:\\ Dimpler}  &\thead{AU15: Lip\\ Corner Depressor}    \\ \hline
\raisebox{-.5\height}{\includegraphics[scale=0.25]{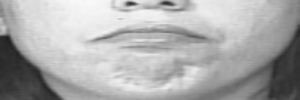}}                                                                 & \raisebox{-.5\height}{\includegraphics[scale=0.25]{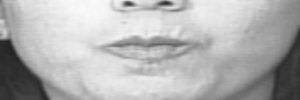}} & \raisebox{-.5\height}{\includegraphics[scale=0.25]{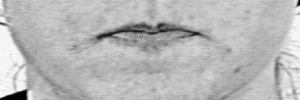}} & \raisebox{-.5\height}{\includegraphics[scale=0.25]{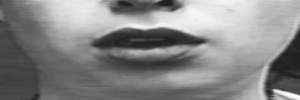}} & \raisebox{-.5\height}{\includegraphics[scale=0.25]{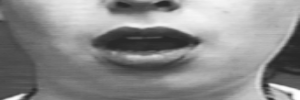}} \\ 
 \thead{AU17:\\ Chin Raiser}&\thead{AU23:\\ Lip Tightener}                                                                &\thead{AU24:\\ Lip Pressor}  &\thead{AU25:\\ Lips Part}  & \thead{AU26:\\ Jaw Drop} \\ \hline
\end{tabular}
\caption{\label{tab:aus}Descriptions and example images for related AUs \cite{FACS1}. }
\end{table*}

\subsection{Datasets}
Our occlusion-robust model is evaluated on two widely-used datasets for AU detection: BP4D \cite{zhang2013high} and DISFA \cite{mavadati2013disfa}. Descriptions and sample images for AUs contained in these datasets are shown in Table \ref{tab:aus}. For both datasets, we evaluated our proposed method using three-fold cross-validation and reported the mean performance values over the folds. For a fair comparison, we adopt the folds split following prior works \cite{shao2018deep, luo2022learning}.

The BP4D dataset \cite{zhang2013high} contains images from 41 young adults (18 male and 23 female) of various ethnicity. Each subject is asked to perform 8 tasks corresponding to different target emotions. There are 328 videos collected, including around 140,000 frames with binary occurrence AU labels (present or absent) on 12 AUs (1, 2, 4, 6, 7, 10, 12, 14, 15, 17, 23, 24). The original resolution of the frames is $1392 \times 1040$ and each contains exactly one front-facing face in the middle of the frame. To avoid the situation that testing data and training data share images from the same person, training and testing partitions in each fold only contain images from different people.

The DISFA \cite{mavadati2013disfa} dataset contains video recordings from 27 subjects (15 males and 12 females) when watching video clips. The dataset contains around 130,000 valid face color images with a resolution of $1024 \times 768$, each of which is with intensity labels on 8 AUs (1, 2, 4, 6, 9, 12, 25, 26). Following prior work, AUs with an intensity equal to or greater than 2 are considered present while others are treated as absent. The same training and testing split strategy as in BP4D is applied to this dataset. This dataset has a more significant AU occurrence imbalance problem, where certain AU could have 5 times more occurrence when compared to the ones with lower occurrence.

\begin{figure}
    \centering
    \includegraphics[scale=0.499]{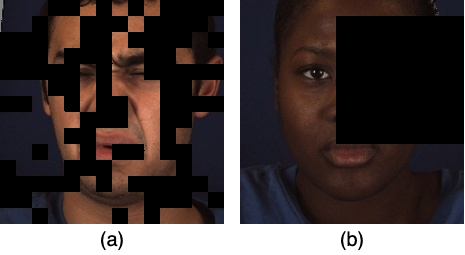}
    \caption{\label{fig:masking_sample} Visualization of two typical facial occlusion conditions for FAU detection: (a) 50\% small-patch occlusion; (b) 30\% large-block occlusion.}

\end{figure}

Due to a lack of real-world annotated FAU occlusion datasets, we simulate the occlusions by masks following ideas from existing face recognition work \cite{qiu2021end2end}. Specifically, we generate two types of masks: small block patches with a size of $16\times 16$ to simulate randomly placed gadgets (e.g., sunglasses, stains), and a relatively large block to simulate large objects e.g., facial masks or palms. In the first type of mask (Figure \ref{fig:masking_sample} (a)), considering that FAU describes very subtle facial muscle movement, we limited the overall masking ratio to around 50\% to avoid covering all FAU-related regions. In the second type (Figure \ref{fig:masking_sample} (b)), the block region is set to be 30\%. It is worth mentioning that our proposed method can be applied to many real-world occlusions by simply adding a pre-processing procedure, i.e., detecting occlusion regions and converting them to binary masks, though obtaining AU annotations for such real-world occluded images remains a challenge currently.

\subsection{Evaluation Metric}
We follow the previous AU detection studies \cite{shao2018deep, jacob2021facial, luo2022learning} and use the F1-score as the metric to evaluate the performance of our approach. F1-score is the harmonic mean of the precision $P$ and the recall $R$, i.e., $F_1 = 2\frac{P\cdot R}{P + R}$, and it is considered a better metric in the case of imbalanced classes (e.g., here negative classes dominate in AU detection). The F1-score for each individual AU is reported, and the average score of all AUs is also computed for comparison purposes.

\subsection{Implementation Details}
For each face image, we performed face alignment through a series of similarity transformations based on the provided face landmarks from both BP4D and DISFA datasets. The transformation is shape-preserving which has no impact on the activation status of AUs. The face alignment gives $256 \times 256$ colored images, from which we perform data argumentation including cropping, horizontal flipping, and color jittering, and obtain $224 \times 224$ images as training inputs. During training, we use AdamW \cite{loshchilov2017decoupled} with $\beta_1=0.9, \beta_2=0.999$ and the weight decay of $5\textrm{e}^{-4}$. For training in the first stage, we choose to use $K=4$ nearest neighbors for both BP4D and DISFA datasets. In both the second stage teacher training and knowledge distillation student training, we give the edge classification loss weight $\lambda=0.01$. 

In knowledge distillation training, we adopt $T=2$ in $L^{edge}_{kd}$ and select the weight $\alpha=1$, $\beta=0.1$ for overall loss and $L_{kd}$ respectively. For teacher model training, we train up to 30 epochs in stage 1 with an initial learning rate of $1\textrm{e}^{-4}$ and $7.5\textrm{e}^{-5}$ on BP4D and DISFA, respectively. We also train 20 epochs for the second stage with an initial learning rate of $1\textrm{e}^{-6}$ and $1\textrm{e}^{-5}$ on BP4D and DISFA datasets, respectively. For knowledge distillation on the student model, 10 epochs of training are used with an initial learning rate of $1\textrm{e}^{-5}$ on both datasets. All the phases of training are done on a single RTX 3090 GPU with a batch size of 48. The initialization of MAE models used in both teacher and student networks are trained on ImageNet\cite{deng2009imagenet}. To fine-tune a pretrained MAE on facial datasets, we set the learning rate on MAE model parameters to 1/100 of the learning rate of rest parameters. The backbone used in the FAU detection module is also pretrained on ImageNet.

\begin{table*}[!htbp]
\centering
\small
\setlength{\tabcolsep}{2.5pt}
\begin{tabular}{c|cccccccccccc|c}
\hline
\multirow{2}{*}{Method}                                                                                          & \multicolumn{12}{c}{AU}                                                                                                                                          & \multirow{2}{*}{Avg.} \\ 
                                                                                                                 & {1} & {2} & {4} & {6} & {7} & {10} & {12} & {14} & {15} & {17} & {23} & {24} &                      \\ \hline
JAA-Net\cite{shao2018deep}                                                                                  & 47.2       & 44.0       & 54.9       & 77.5       & 74.6       & 84.0        & 86.9        & 61.9        & 43.6        & 60.3        & 42.7        & 41.9        & 60.0                 \\
AU-GCN\cite{liu2020relation}                                                                                   & 46.8       & 38.5       & 60.1       & 80.1       & 79.5       & 84.8        & 88.0        & 67.3        & 52.0        & 63.2        & 40.9        & 52.8        & 62.8                 \\

UGN-B\cite{song2021uncertain}                                                                                    & 54.2       & 46.4       & 56.8       & 76.2       & 76.7       & 82.4        & 86.1        & 64.7        & 51.2        & 63.1        & 48.5        & 53.6        & 63.3                 \\
SEV-Net\cite{yang2021exploiting}                                                                                  & \underline{58.2}       & \underline{50.4}       & 58.3       & \underline{81.9}       & 73.9       & \underline{87.8}        & 87.5        & 61.6        & 52.6        & 62.2        & 44.6        & 47.6        & 63.9                 \\
ME-GraphAU\cite{luo2022learning}                                                                               & 52.7       & 44.3       & \underline{60.9}       & 79.9       & \underline{80.1}       & 85.3        & \underline{89.2}        & \underline{69.4}        & \underline{55.4}        & \underline{64.4}        & \underline{49.8}        & 55.1        & \underline{65.5}                 \\ \hline
\begin{tabular}[c]{@{}c@{}}ME-GraphAU:\\  50\% random mask\end{tabular}                   & 15.12      & 0.50       & 35.28      & 4.16       & 23.44      & 57.58       & 74.23       & 65.30       & 24.44       & 16.53       & 16.12       & 36.11       & 30.73                \\  
\begin{tabular}[c]{@{}c@{}}ME-GraphAU: \\ 50\% random mask\\ reconstructed\end{tabular} & 41.44      & 47.40      & 45.41      & 69.90      & 66.02      & 79.47       & 86.74       & 62.88       & 32.11       & 57.91       & 34.37       & 27.98       & 54.30                \\ \hdashline
\begin{tabular}[c]{@{}c@{}}\textbf{Ours (teacher):} \\ \textbf{50\% random mask} \end{tabular}             & \textbf{51.55}      & \textbf{46.40}      & 55.84      & \textbf{78.61}      & 78.33      & \textbf{83.11}       & 88.16       & \textbf{66.72}       & 50.08       & \textbf{63.81}       & \textbf{50.28}       & 47.63       & \textbf{63.38}                \\ 
\begin{tabular}[c]{@{}c@{}} \textbf{Ours (student):} \\ \textbf{50\% random mask} \end{tabular}             & 50.61      & 43.21      & \textbf{56.56}      & 77.97      & \textbf{78.65}      & 82.95       & \textbf{87.72}       & 66.43       & 47.62       & 63.66       & 45.71       & \textbf{49.67}       & 62.56                \\ \hline
\begin{tabular}[c]{@{}c@{}}ME-GraphAU: \\ 30\% block mask\end{tabular}                     & 40.19      & 42.27      & 46.52      & 66.41      & 52.76      & 79.39       & 78.50       & 60.77       & 30.13       & 53.11       & 29.73       & 29.92       & 50.81                \\
\begin{tabular}[c]{@{}c@{}}ME-GraphAU: \\ 30\% block mask\\ reconstructed\end{tabular}     & 42.00      & 45.21      & 49.59      & 72.63      & 68.12      & 76.85       & 83.68       & 55.62       & 26.20       & 54.01       & 24.86       & 27.39       & 52.18                \\ \hdashline
\begin{tabular}[c]{@{}c@{}}\textbf{Ours (teacher):}\\ \textbf{30\% block mask}\end{tabular}                 & 47.15      & 43.37      & 55.90      & {77.04}      & \doubleunderline{77.58}      & \doubleunderline{81.57}       & \doubleunderline{87.09}       & \doubleunderline{67.62}       & \doubleunderline{51.77}       & \doubleunderline{62.02}       & \doubleunderline{44.64}       & \doubleunderline{49.37}       & \doubleunderline{62.09}                \\
\begin{tabular}[c]{@{}c@{}} \textbf{Ours (student):}\\ \textbf{30\% block mask}\end{tabular}                 & \doubleunderline{48.37}      & \doubleunderline{44.30}      & \doubleunderline{57.46}      & 76.73      & 77.35      & 81.51       & 85.09       & 66.36       & 48.56       & 61.23       & 42.87       & 45.09       & 61.24\\\hline              
\end{tabular}
\caption{\label{tab:BP4D} Comparison of F1-scores (\%) for 12 AUs on BP4D dataset. The top section of the table are results from different baseline models on occlusion-free images. The middle and bottom sections contain results from the SOTA method ME-GraphAU\cite{luo2022learning} and our method on images with different types of occlusions. The best results of each section are highlighted with underline, bold font and double underline,  respectively. }
\end{table*}

\begin{table*}[!htbp]
\centering
\small
\begin{tabular}{c|cccccccc|c}
\hline
\multirow{2}{*}{Method}                                                               & \multicolumn{8}{c}{AU}                                                                                                                                                                & \multirow{2}{*}{Avg.} \\
                                                                                      & {1}                    & {2}                    & {4}                    & {6}                    & {9}                    & {12}                   & {25}                   & {26}                   &                       \\ \hline
JAA-Net\cite{shao2018deep}                                                                               & 43.7                 & 46.2                 & 56.0                 & 41.4                 & 44.7                 & 69.6                 & 88.3                 & 58.4                 & 56.0                  \\
AU-GCN\cite{liu2020relation}                                                                                & 32.3                 & 19.5                 & 55.7                 & \underline{57.0}                 & 61.4                 & 62.7                 & 90.9                 & \underline{60.0}                 & 55.0                  \\
UGN-B\cite{song2021uncertain}                                                                                 & 43.3                 & 48.1                 & 63.4                 & 49.5                 & 48.2                 & 72.9                 & 90.8                 & 59.0                 & 60.0                  \\
SEV-Net\cite{yang2021exploiting}                                                                               & \underline{55.3}                 & \underline{53.1}                 & 61.5                 & 53.6                 & 38.2                 & 71.6                 & \underline{95.7}                 & 41.5                 & 58.8                  \\
ME-GraphAU\cite{luo2022learning}                                                                            & 54.6                 & 47.1                 & \underline{72.9}                 & 54.0                 & 55.7                 & 76.7                 & 91.1                 & 53.0                 & \underline{63.1}                  \\ \hline
\begin{tabular}[c]{@{}c@{}}ME-GraphAU:\\ 50\% random mask\end{tabular}                 & 25.29                & 22.91                & 47.72                & 27.83                & 25.44                & 50.63                & 69.53                & 31.30                & 37.58                 \\
\begin{tabular}[c]{@{}c@{}}ME-GraphAU:\\ 50\% random mask\\ reconstructed\end{tabular} & 40.82                & 34.37                & 61.64                & 31.85                & \textbf{44.19}                & 73.09                & 89.84                & \textbf{61.95}                & 54.72                 \\  \hdashline
\begin{tabular}[c]{@{}c@{}}\textbf{Ours (teacher):}\\ \textbf{50\% random mask}\end{tabular}             & \textbf{60.58}                & 48.64                & 60.48                & 46.60                & 44.08                & \textbf{73.43}                & \textbf{91.14}                & 59.96                & 60.62                 \\
\begin{tabular}[c]{@{}c@{}} \textbf{Ours (student):}\\ \textbf{50\% random mask}\end{tabular}             & 53.83                & \textbf{54.31}                & \textbf{67.02}                & \textbf{49.15}                & 41.22                & 73.19                & 91.06                & 60.19                & \textbf{61.25}                 \\ \hline
\begin{tabular}[c]{@{}c@{}} {ME-GraphAU:}\\ {30\% block mask}\end{tabular}                 & 35.66                & 23.47                & 54.87                & 26.42                & 28.13                & 57.04                & 67.84                & 42.66                & 42.01                 \\
\begin{tabular}[c]{@{}c@{}}ME-GraphAU:\\ 30\% block mask\\ reconstructed\end{tabular} & 40.03                & 28.92                & 55.91                & 26.03                & 35.00                & 66.01                & 74.48                & \doubleunderline{49.31}                & 47.00                 \\  \hdashline
\begin{tabular}[c]{@{}c@{}}\textbf{Ours (teacher):}\\ \textbf{30\% random mask}\end{tabular}             & 48.43                & 42.04                & \doubleunderline{61.69}                & \doubleunderline{48.13}                & \doubleunderline{46.35}                & \doubleunderline{69.98}                & \doubleunderline{83.41}                & 48.98                & \doubleunderline{56.13}                 \\
\begin{tabular}[c]{@{}c@{}}\textbf{Ours (student):}\\ \textbf{30\% random mask}\end{tabular}             & \doubleunderline{50.66}                & \doubleunderline{43.20}                & 59.64                & 47.62                & 38.94                & 68.52                & 81.53                & 44.88                & 54.37                 \\\hline
\end{tabular}
\caption{\label{tab:DISFA} Comparison of F1-scores (in\%) for 8 AUs on DISFA dataset. The top section of the table are results from different baseline models on occlusion-free images. The middle and bottom sections contain results from the SOTA method ME-GraphAU \cite{luo2022learning} and our method on images with different types of occlusions. The best results of each section are highlighted with underline, bold font and double underline,  respectively.}
\end{table*}

\subsection{Experimental Results}
In this section, we compare our results with several state-of-the-art methods on both datasets under a few different settings. Table \ref{tab:BP4D} reports the occurrence detection results of 12 AUs on BP4D dataset in terms of F1-score. The top section of the table contains results under the occlusion-free conditions from different recent baseline methods (i.e. JAA-Net\cite{shao2018deep}, AU-GCN\cite{liu2020relation}, SEV-Net\cite{yang2021exploiting} and ME-GraphAU\cite{luo2022learning}), which are reported to have better performance than representative earlier methods including JPML\cite{zhao2015joint}, DRML\cite{zhao2016deep}, EAC-Net\cite{eacnet2017} and DSIN \cite{corneanu2018deep} etc. The bottom section contains experimental results under various occlusion conditions. As we can see from the table, even with 30\% to 50\% occlusions, both the teacher and student models we proposed can achieve the same level of performance as other models under occlusion-free conditions.  As mentioned in the previous section, the state-of-the-art models trained on regular high-quality images degenerated significantly when we introduce different occlusion conditions. E.g., for ME-GraphAU, the F1-score drops from 65.5\% to 30.73\% and 50.81\% on 50\% and 30\% occlusion respectively. Reconstructing occluded images using the ImageNet pre-trained MAE does help in the AU detection performance by filling in the missing information, especially for the high percentage sparse occlusion case where the F1-score is increased from 30.73\% to 54.3\% for ME-GraphAU. However, this accuracy is still over 10\% away from its original performance on occlusion-free images. By contrast, our proposed models are forced to learn rich AU features and reliable AU relationships from only visible areas. And the performance of AU detection under 50\% random occlusion is boosted to 63.38\% and 62.56\% in our reconstruction-based teacher model and the efficient student model respectively. In the case of 30\% occlusion, our proposed models again significantly improve the performance by 10\% over the ME-GraphAU model, achieving non-occlusion level performance with an F1-score of 62.09\% and 61.24\% when using the teacher and student models respectively. 

In Table \ref{tab:DISFA}, we show experimental results on the DISFA dataset using the same 30\% block and 50\% random occlusion configurations. From the table, our proposed models again improve the performance of occurrence detection on 8 AUs under the occluded conditions by a large margin. Our proposed models can achieve 60.62\% and 61.25\% under the 50\% random occlusion condition, which is even better than most other methods under the occlusion-free condition. For the 50\% random occlusion condition, the student model using latent space features achieves a better result than the reconstruction-based teacher model in our experiments. Unlike the BP4D dataset, the data variance of DISFA is much smaller. The reconstruction module could suffer from overfitting the training data by reconstructing similar images again and again because of the small number of unique images from the DISFA dataset. The overfitting phenomenon is further exaggerated on the block masking experiments on DISFA dataset where we found some occluded regions are reconstructed with blocks from other faces from the training dataset. Such an overfitting problem could cause huge noise in supervision and thus limit the ability for efficient learning. As seen from the 30\% block masking experimental results, though better than the state-of-art ME-GraphAU, the performance gains of our proposed models are limited.

By comparing the results from the 30\% block occlusion condition and the 50\% random occlusion condition, we can see that the occlusion condition has a very pronounced effect on the relationship and feature learning. Although the single large block type of occlusion has a smaller coverage, it brings more challenges in AU detection. Normally, under the non-occlusion condition, the activation of each AU is mostly determined by local features with the aid of the relationship between AUs. Under the large block occlusion, all nearby regions could be occluded and the models are forced to use only features from the far-away region and inter-relationship information between occluded and visible regions to do the inference. We can see that, for the BP4D dataset, the state-of-the-art model ME-GraphAU has only a 2\% performance gain from reconstruction under the large block occlusion setting, while a 24\% gain under the random occlusion setting. Our models also have lower performance gains under the large block occlusion than under the random occlusion, because limited local information can be extracted for reconstruction as well as for AU detection.

Based on the above experiments, we observe that our proposed models can significantly improve the AU detection performance under various heavy occlusions, achieving comparable performances with other state-of-the-art models under occlusion-free conditions.

\subsection{Robustness Assessment}

\begin{figure}[!htbp]
\centering
\includegraphics[scale=0.5]{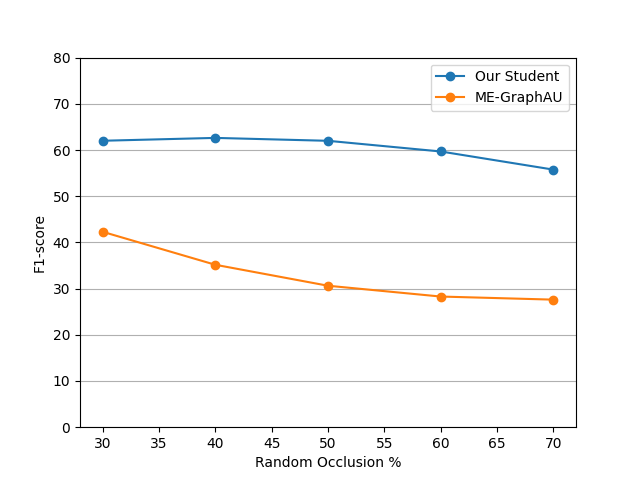}
\caption{\label{fig:f1_vs_occlusion_ratio} The trend of F1-score over an increasing percentage of random occlusion (\%) on BP4D dataset when using ME-GraphAU\cite{luo2022learning} and our student model.}
\end{figure}

In this section, we aim to show the robustness of our proposed model against different levels of occlusions. In this study, we choose 1 of the 3 folds from the BP4D dataset. Our model is trained with 50\% small patch random occlusions. We reported in Fig. \ref{fig:f1_vs_occlusion_ratio} the F1-score comparisons between our model and the SOTA model ME-GraphAU on the testing dataset under various levels of occlusions. As we can see, the performance of our proposed model in terms of F1-score is relatively stable under occlusion rations from 30\% to 70\%, confirming that our model is robust against different occlusions. In particular, even on face images under 70\% occlusion conditions, our model can still achieve an F1-score as high as 55.8\%, significantly outperforming the SOTA model ME-GraphAU.

\subsection{Computation Efficiency Comparison}
In addition to performance comparisons, this section compares the computation efficiency between our proposed model and the state-of-the-art FAU recognition model. In Table \ref{tab:comp}, we show the computational complexity and the number of parameters of ME-GraphAU, our teacher and student models. We can note that, although our reconstruction-based teacher model generally achieves the best performances under occlusion settings on BP4D and DISFA datasets, it also requires a high computational cost (50\% more than that of ME-GraphAU) and a larger number of model parameters (100\% more than that of ME-GraphAU). While our proposed student model, maintaining high performance under heavy occlusion conditions, has slightly less computational complexity as well as model size when compared with ME-GraphAU. 


\begin{table}[]
\centering
\begin{adjustbox}{width=9cm}
\begin{tabular}{|c|c|c|}
\hline
Model      & \begin{tabular}[c]{@{}c@{}}Computational Complexity \end{tabular} & No. of Parameters \\ \hline
ME-GraphAU\cite{luo2022learning} & 21.28 GMACs                                                          & 94.38 M       \\ \hline
Teacher (ours)   & 30.29  GMACs                                                        & 201.47 M      \\ \hline
Student (ours)   & \textbf{18.63 GMACs}                                                         & \textbf{91.26 M}      \\ \hline
\end{tabular}
\end{adjustbox}
\caption{\label{tab:comp} Comparisons of computational efficiency between different model backbones. Best performances have been highlighted in bold font. }
\end{table}

\section{Conclusion}
\label{sec:conclusion}
This work proposed a novel framework for facial action unit recognition under heavy occlusion conditions. Our reconstruction-based model, taking advantage of masked image modeling, is robust against heavy occlusions by learning the rich FAU-related features only from the visible parts of the facial image. The proposed models incorporate graph edge feature learning to further mitigate the influence of occlusion by shifting the focus from local feature learning to AU relationship learning. Further, we improve the efficiency of our model by transferring the latent space features of the masked autoencoder to FAU features by performing both edge-level and node-level knowledge distillation. The results on two commonly used datasets demonstrate that the proposed models under the 50\% random occlusion can achieve comparable results with the state-of-the-art method under occlusion-free conditions. Our proposed occlusion-robust facial action unit recognition methods are modular by design and can be easily extended to other similar problems to enhance the robustness under heavy occlusions.

\bibliographystyle{elsarticle-num} 
\bibliography{elsarticle}

\begin{thebibliography}{10}
\expandafter\ifx\csname url\endcsname\relax
  \def\url#1{\texttt{#1}}\fi
\expandafter\ifx\csname urlprefix\endcsname\relax\def\urlprefix{URL }\fi
\expandafter\ifx\csname href\endcsname\relax
  \def\href#1#2{#2} \def\path#1{#1}\fi

\bibitem{FACS1}
P.~Ekman, W.~V. Friesen, Facial action coding system: a technique for the
  measurement of facial movement, 1978.

\bibitem{fau_survey_2020}
R.~Zhi, M.~Liu, D.~Zhang, \href{https://doi.org/10.1007/s00371-019-01707-5}{A
  comprehensive survey on automatic facial action unit analysis}, Vis. Comput.
  36~(5) (2020) 1067–1093.
\newblock \href {https://doi.org/10.1007/s00371-019-01707-5}
  {\path{doi:10.1007/s00371-019-01707-5}}.
\newline\urlprefix\url{https://doi.org/10.1007/s00371-019-01707-5}

\bibitem{zhao2015joint}
K.~Zhao, W.-S. Chu, F.~De~la Torre, J.~F. Cohn, H.~Zhang, Joint patch and
  multi-label learning for facial action unit detection, in: Proceedings of the
  IEEE Conference on Computer Vision and Pattern Recognition, 2015, pp.
  2207--2216.

\bibitem{jaiswal2016deep}
S.~Jaiswal, M.~Valstar, Deep learning the dynamic appearance and shape of
  facial action units, in: 2016 IEEE winter conference on applications of
  computer vision (WACV), IEEE, 2016, pp. 1--8.

\bibitem{li2017eac}
W.~Li, F.~Abtahi, Z.~Zhu, L.~Yin, Eac-net: A region-based deep enhancing and
  cropping approach for facial action unit detection, in: 2017 12th IEEE
  International Conference on Automatic Face \& Gesture Recognition (FG 2017),
  IEEE, 2017, pp. 103--110.

\bibitem{eacnet2017}
W.~Li, F.~Abtahi, Z.~Zhu, L.~Yin, Eac-net: A region-based deep enhancing and
  cropping approach for facial action unit detection, in: 2017 12th IEEE
  International Conference on Automatic Face \& Gesture Recognition (FG 2017),
  2017, pp. 103--110.
\newblock \href {https://doi.org/10.1109/FG.2017.136}
  {\path{doi:10.1109/FG.2017.136}}.

\bibitem{corneanu2018deep}
C.~Corneanu, M.~Madadi, S.~Escalera, Deep structure inference network for
  facial action unit recognition, in: Proceedings of the european conference on
  computer vision (ECCV), 2018, pp. 298--313.

\bibitem{cohn1999automated}
J.~F. Cohn, A.~J. Zlochower, J.~Lien, T.~Kanade, Automated face analysis by
  feature point tracking has high concurrent validity with manual facs coding,
  Psychophysiology 36~(1) (1999) 35--43.

\bibitem{whitehill2006haar}
J.~Whitehill, C.~W. Omlin, Haar features for facs au recognition, in: 7th
  international conference on automatic face and gesture recognition (FGR06),
  IEEE, 2006, pp. 5--pp.

\bibitem{liu2020relation}
Z.~Liu, J.~Dong, C.~Zhang, L.~Wang, J.~Dang, Relation modeling with graph
  convolutional networks for facial action unit detection, in: International
  Conference on Multimedia Modeling, Springer, 2020, pp. 489--501.

\bibitem{shao2018deep}
Z.~Shao, Z.~Liu, J.~Cai, L.~Ma, Deep adaptive attention for joint facial action
  unit detection and face alignment, in: Proceedings of the European conference
  on computer vision (ECCV), 2018, pp. 705--720.

\bibitem{luo2022learning}
C.~Luo, S.~Song, W.~Xie, L.~Shen, H.~Gunes, Learning multi-dimensional edge
  feature-based au relation graph for facial action unit recognition, arXiv
  preprint arXiv:2205.01782 (2022).

\bibitem{song2019occlusion}
L.~Song, D.~Gong, Z.~Li, C.~Liu, W.~Liu, Occlusion robust face recognition
  based on mask learning with pairwise differential siamese network, in:
  Proceedings of the IEEE/CVF International Conference on Computer Vision,
  2019, pp. 773--782.

\bibitem{zhang2022learning}
Y.~Zhang, X.~Wang, M.~S. Shakeel, H.~Wan, W.~Kang, Learning upper patch
  attention using dual-branch training strategy for masked face recognition,
  Pattern Recognition 126 (2022) 108522.

\bibitem{qiu2021end2end}
H.~Qiu, D.~Gong, Z.~Li, W.~Liu, D.~Tao, End2end occluded face recognition by
  masking corrupted features, IEEE Transactions on Pattern Analysis and Machine
  Intelligence (2021).

\bibitem{zeng2021survey}
D.~Zeng, R.~Veldhuis, L.~Spreeuwers, A survey of face recognition techniques
  under occlusion, IET biometrics 10~(6) (2021) 581--606.

\bibitem{song2021uncertain}
T.~Song, L.~Chen, W.~Zheng, Q.~Ji, Uncertain graph neural networks for facial
  action unit detection, in: Proceedings of the AAAI Conference on Artificial
  Intelligence, Vol.~35, 2021, pp. 5993--6001.

\bibitem{rani2018systematic}
M.~Rani, S.~B. Dhok, R.~B. Deshmukh, A systematic review of compressive
  sensing: Concepts, implementations and applications, IEEE Access 6 (2018)
  4875--4894.

\bibitem{kulkarni2016reconnet}
K.~Kulkarni, S.~Lohit, P.~Turaga, R.~Kerviche, A.~Ashok, Reconnet:
  Non-iterative reconstruction of images from compressively sensed
  measurements, in: Proceedings of the IEEE Conference on Computer Vision and
  Pattern Recognition, 2016, pp. 449--458.

\bibitem{wang2018revhashnet}
Y.~Wang, H.~Palangi, Z.~J. Wang, H.~Wang, Revhashnet: Perceptually de-hashing
  real-valued image hashes for similarity retrieval, Signal processing: Image
  communication 68 (2018) 68--75.

\bibitem{he2022masked}
K.~He, X.~Chen, S.~Xie, Y.~Li, P.~Doll{\'a}r, R.~Girshick, Masked autoencoders
  are scalable vision learners, in: Proceedings of the IEEE/CVF Conference on
  Computer Vision and Pattern Recognition, 2022, pp. 16000--16009.

\bibitem{zhao2016deep}
K.~Zhao, W.-S. Chu, H.~Zhang, Deep region and multi-label learning for facial
  action unit detection, in: Proceedings of the IEEE conference on computer
  vision and pattern recognition, 2016, pp. 3391--3399.

\bibitem{jacob2021facial}
G.~M. Jacob, B.~Stenger, Facial action unit detection with transformers, in:
  Proceedings of the IEEE/CVF Conference on Computer Vision and Pattern
  Recognition, 2021, pp. 7680--7689.

\bibitem{yang2021exploiting}
H.~Yang, L.~Yin, Y.~Zhou, J.~Gu, Exploiting semantic embedding and visual
  feature for facial action unit detection, in: Proceedings of the IEEE/CVF
  Conference on Computer Vision and Pattern Recognition, 2021, pp.
  10482--10491.

\bibitem{bao2022beit}
H.~Bao, L.~Dong, S.~Piao, F.~Wei,
  \href{https://openreview.net/forum?id=p-BhZSz59o4}{{BE}it: {BERT}
  pre-training of image transformers}, in: International Conference on Learning
  Representations, 2022.
\newline\urlprefix\url{https://openreview.net/forum?id=p-BhZSz59o4}

\bibitem{Xie_2022_CVPR}
Z.~Xie, Z.~Zhang, Y.~Cao, Y.~Lin, J.~Bao, Z.~Yao, Q.~Dai, H.~Hu, Simmim: A
  simple framework for masked image modeling, in: Proceedings of the IEEE/CVF
  Conference on Computer Vision and Pattern Recognition (CVPR), 2022, pp.
  9653--9663.

\bibitem{gou2021knowledge}
J.~Gou, B.~Yu, S.~J. Maybank, D.~Tao, Knowledge distillation: A survey,
  International Journal of Computer Vision 129~(6) (2021) 1789--1819.

\bibitem{ding2023distilling}
X.~Ding, Y.~Wang, Z.~Xu, Z.~J. Wang, W.~J. Welch, Distilling and transferring
  knowledge via cgan-generated samples for image classification and regression,
  Expert Systems with Applications 213 (2023) 119060.

\bibitem{hinton2015distilling}
G.~Hinton, O.~Vinyals, J.~Dean, et~al., Distilling the knowledge in a neural
  network, arXiv preprint arXiv:1503.02531 2~(7) (2015).

\bibitem{romero2014fitnets}
A.~Romero, N.~Ballas, S.~E. Kahou, A.~Chassang, C.~Gatta, Y.~Bengio, Fit{N}et:
  Hints for thin deep nets, in: International Conference on Learning
  Representations, 2015.

\bibitem{xu2020knowledge}
G.~Xu, Z.~Liu, X.~Li, C.~C. Loy, Knowledge distillation meets self-supervision,
  in: European Conference on Computer Vision, Springer, 2020, pp. 588--604.

\bibitem{wang2022ssd}
Y.~Wang, Y.~Wang, J.~Cai, T.~K. Lee, C.~Miao, Z.~J. Wang, Ssd-kd: A
  self-supervised diverse knowledge distillation method for lightweight skin
  lesion classification using dermoscopic images, Medical Image Analysis (2022)
  102693.

\bibitem{liu2018multi}
Y.~Liu, L.~Sheng, J.~Shao, J.~Yan, S.~Xiang, C.~Pan, Multi-label image
  classification via knowledge distillation from weakly-supervised detection,
  in: Proceedings of the 26th ACM international conference on Multimedia, 2018,
  pp. 700--708.

\bibitem{xu2022boosting}
J.~Xu, S.~Huang, F.~Zhou, L.~Huangfu, D.~Zeng, B.~Liu, Boosting multi-label
  image classification with complementary parallel self-distillation, arXiv
  preprint arXiv:2205.10986 (2022).

\bibitem{song2021handling}
L.~Song, J.~Wu, M.~Yang, Q.~Zhang, Y.~Li, J.~Yuan, Handling difficult labels
  for multi-label image classification via uncertainty distillation, in:
  Proceedings of the 29th ACM International Conference on Multimedia, 2021, pp.
  2410--2419.

\bibitem{vit}
A.~Dosovitskiy, L.~Beyer, A.~Kolesnikov, D.~Weissenborn, X.~Zhai,
  T.~Unterthiner, M.~Dehghani, M.~Minderer, G.~Heigold, S.~Gelly, et~al., An
  image is worth 16x16 words: Transformers for image recognition at scale,
  arXiv preprint arXiv:2010.11929 (2020).

\bibitem{ben2020asymmetric}
E.~Ben-Baruch, T.~Ridnik, N.~Zamir, A.~Noy, I.~Friedman, M.~Protter,
  L.~Zelnik-Manor, Asymmetric loss for multi-label classification, arXiv
  preprint arXiv:2009.14119 (2020).

\bibitem{bresson2017residual}
X.~Bresson, T.~Laurent, Residual gated graph convnets, arXiv preprint
  arXiv:1711.07553 (2017).

\bibitem{zhang2013high}
X.~Zhang, L.~Yin, J.~F. Cohn, S.~Canavan, M.~Reale, A.~Horowitz, P.~Liu, A
  high-resolution spontaneous 3d dynamic facial expression database, in: 2013
  10th IEEE international conference and workshops on automatic face and
  gesture recognition (FG), IEEE, 2013, pp. 1--6.

\bibitem{mavadati2013disfa}
S.~M. Mavadati, M.~H. Mahoor, K.~Bartlett, P.~Trinh, J.~F. Cohn, Disfa: A
  spontaneous facial action intensity database, IEEE Transactions on Affective
  Computing 4~(2) (2013) 151--160.

\bibitem{loshchilov2017decoupled}
I.~Loshchilov, F.~Hutter, Decoupled weight decay regularization, arXiv preprint
  arXiv:1711.05101 (2017).

\bibitem{deng2009imagenet}
J.~Deng, W.~Dong, R.~Socher, L.-J. Li, K.~Li, L.~Fei-Fei, Imagenet: A
  large-scale hierarchical image database, in: 2009 IEEE conference on computer
  vision and pattern recognition, IEEE, 2009, pp. 248--255.

\end{thebibliography}

\end{document}